\documentclass[onefignum,onetabnum]{siamonline190516}

\newcommand{\grad}{\ensuremath{\nabla}}

\newcommand{\bfA}{{\bf A}}

\newcommand{\bfC}{{\bf C}}
\newcommand{\bfD}{{\bf D}}

\newcommand{\bfI}{{\bf I}}

\newcommand{\bfK}{{\bf K}}

\newcommand{\bfP}{{\bf P}}
\newcommand{\bfQ}{{\bf Q}}

\newcommand{\bfT}{{\bf T}}

\newcommand{\bfW}{{\bf W}}
\newcommand{\bfX}{{\bf X}}
\newcommand{\bfY}{{\bf Y}}
\newcommand{\bfZ}{{\bf Z}}

\newcommand{\bfb}{{\bf b}}

\newcommand{\bftheta}{{\boldsymbol \theta}}

\renewcommand{\theequation}{\arabic{section}.\arabic{equation}}

\definecolor{mycolor}{rgb}{0.3,0.44,1}
\definecolor{targetcolor}{rgb}{0.5,0.64,1}

\newcommand{\bfzeta}{{\boldsymbol \zeta}}

\usepackage{tikz}
\usepackage{lipsum}
\usepackage{amsfonts}
\usepackage{graphicx}
\usepackage{epstopdf}
\usepackage{algorithmic}
\usepackage{subcaption}
\ifpdf
  \DeclareGraphicsExtensions{.eps,.pdf,.png,.jpg}
\else
  \DeclareGraphicsExtensions{.eps}
\fi

\usepackage{enumitem}
\setlist[enumerate]{leftmargin=.5in}
\setlist[itemize]{leftmargin=.5in}

\newsiamremark{remark}{Remark}
\newsiamremark{hypothesis}{Hypothesis}
\crefname{hypothesis}{Hypothesis}{Hypotheses}
\newsiamthm{claim}{Claim}

\headers{Fully Hyperbolic Convolutional Neural Networks}{K. Lensink, B. Peters, E. Haber}

\title{Fully Hyperbolic Convolutional Neural Networks}

\author{Keegan Lensink\thanks{The University of British Columbia and Xtract AI, Vancouver, Canada. 
    \email{klensink@eoas.ubc.ca}, \email{ehaber@eoas.ubc.ca}}
\and Bas Peters\thanks{Computational Geosciences Inc., \email{bas@compgeoinc.com}}
\and Eldad Haber\footnotemark[1]}

\usepackage{amsopn}

\begin{document}

\maketitle

\begin{abstract}
Convolutional Neural Networks (CNN) have recently seen tremendous success in various computer vision tasks. 
However, their application to problems with high dimensional input and output, such as high-resolution image and video segmentation or 3D medical imaging, has been limited by various factors. 
Primarily, in the training stage, it is necessary to store network activations for back propagation. In these settings, the memory requirements associated with storing activations can exceed what is feasible with current hardware, especially for problems in 3D. 
Motivated by the propagation of signals over physical networks, that are governed by the hyperbolic Telegraph equation, in this work we introduce a fully conservative hyperbolic network for problems with high dimensional input and output. 
We introduce a coarsening operation that allows completely reversible CNNs by using a learnable Discrete Wavelet Transform and its inverse to both coarsen and interpolate the network state and change the number of channels. 
We show that fully reversible networks are able to achieve results comparable to the state of the art in 4D time-lapse hyper spectral image segmentation and full 3D video segmentation, with a much lower memory footprint that is a constant independent of the network depth. We also extend the use of such networks to Variational Auto Encoders with high resolution input and output.
\end{abstract}



\section{Introduction}
\label{sec:intro}

Deep Convolutional Neural Networks have recently solved some very challenging problems in computer vision ranging from image classification, segmentation, deblurring, shape from shading, and more \cite{krizhevsky2012imagenet,UNET2015,Tao2018Deblurring,bengio2009learning,lecun2015deep,Goodfellow-et-al-2016,Hammernik_2017,AVENDI2016108}. 
The recent success of neural networks can in part be attributed to the massive amount of data that is being collected allowing the training of complex models with hundreds of millions of parameters. 
Despite recent advances in network architectures permitting deeper and more stable networks we continue to face a number of challenges that current algorithms have yet to solve. 
As we push the scale at which deep learning is applied, computational time and memory requirements are quickly exceeding what is possible with current hardware. 
Memory is especially an issue with deep networks or when the size of the input and output is large and in high dimensions. 
Examples include semantic segmentation and image deblurring or cases where 3D convolutions are required such as medical imaging, seismic applications, and video processing. 
In these cases, memory constraints are a limiting factor when dealing with large scale problems in reasonable computational time and resources. 
In the training phase, all previous states of the network, i.e. its activations, must be stored for backpropagation. 
For these problems, the memory requirements are quickly over-reaching our ability to store the activations of the network.

Beyond just the obvious implications of working with large scale data, such as videos, network depth and width are a significant factor in the memory footprint of a network. 
For fixed width networks, the depth of the network allows us to obtain more nonlinear models and obtain more expressive networks \cite{Hanin2017DNNApprox}. 
Moreover, for problems in vision, networks depth plays another important role, since convolutions are local operations information propagates a fixed distance each layer. 
This implies that the output is determined with information available from a limited patch of the input, the size of which is determined by the number of layers and the width of the convolution kernel. 
The size of this patch is know as the receptive field \cite{NIPS2016_6203}. 
By coarsening the image, the receptive field of the network grows and allows learning non-local relations. 
However, coarsening the image comes with the price of reducing the resolution of the final output. For problems such as classification this is not an issue as we desire a reduction in dimensionality, however for problems with high dimensional output, e.g. semantic segmentation, high resolution output is required. 
In these cases, the image is still coarsened to achieve the desired receptive field, however interpolation is used to regain resolution. 
A consequence of this is that fine image details or high frequency content is typically missing since the coarsening and subsequent interpolation is not conservative. 
In order to re-introduce high frequency information long skip connections are used to connect network states before coarsening with states after interpolation, such as in the popular U-Net architecture \cite{UNET2015}, or features from different resolution scales are combined in Stacked U-Nets \cite{SUnets}. 
Width allows the network to learn more features, increasing the capability of the network to learn. 
However, the width comes at considerable price in terms of memory and computational effort. 
While it is clear that network depth and width are a significant factor in the memory footprint, they are critical to the network's success.

In this work we introduce a new network architecture that addresses all the above difficulties.
The network is {\em fully conservative and reversible}, that is, it can propagate forward and backward without any loss, even though resolution is changed. 
The network communicates in a reversible way between scales, and therefore does not require long skip connections.
Our formulation is motivated by drawing a connection to the propagation of signals over physical networks and hyperbolic Partial Differential Equations. 
In physical networks, such as biological nets, signals can propagate in both directions. 
Indeed, a continuous formulation of a network involves the Telegraph equation (see \cite{Zhou2018Telegraph} and references within) which, upon discretization leads to a different formulation than the canonical ResNet. 
Similar to signal propagation in physical networks, our propagation is fully conservative.
Conservation implies that, although our network has some similarities to the structure of a ResNet, when we coarsen the image we {\em do not lose information} and we can {\em exactly} recover any previous state of the network. 
This means that in the training phase we do not need to store the activations, and the memory footprint is independent of the network's depth. 
We facilitate networks with complete receptive fields, even with high resolution input, by removing the memory limitations related to deep networks.

\bigskip

Given the design of our network we can use it for tasks such as semantic segmentation moreover, the ability to control the information content carried forward (and backward) by our network enable us to use our network as a variational auto-encoder. We use our network to carry the information forward to the latent space without loss, truncate the information in the latent space and only then return it to the original space. This is similar to the Truncated SVD where a linear transformation takes the information to a latent space and is reduced in the latent space.

The rest of the paper is structured as follows. 
In Section~\ref{contribution} we review prior work and clearly state our contributions. 
In \ref{buildingBlocks} we review reversible architectures, wavelets, discuss how one can combine low resolution channels in order to obtain a single high resolution image and introduce our new network architecture. 
In Section \ref{AE} we show how our network can be used to generate a variational auto-encoder that is similar in concept to the Truncated SVD.
In Section~\ref{sec:opt} we discuss the reversible training algorithm.
In Section~\ref{experiments} we experiment with our network on a number of problems and show its efficiency and finally, in Section \ref{conclusions} we summarize the paper.

\section{Contributions and prior work}
\label{contribution}

\begin{figure}
    \centering
        \includegraphics[width=0.70\textwidth]{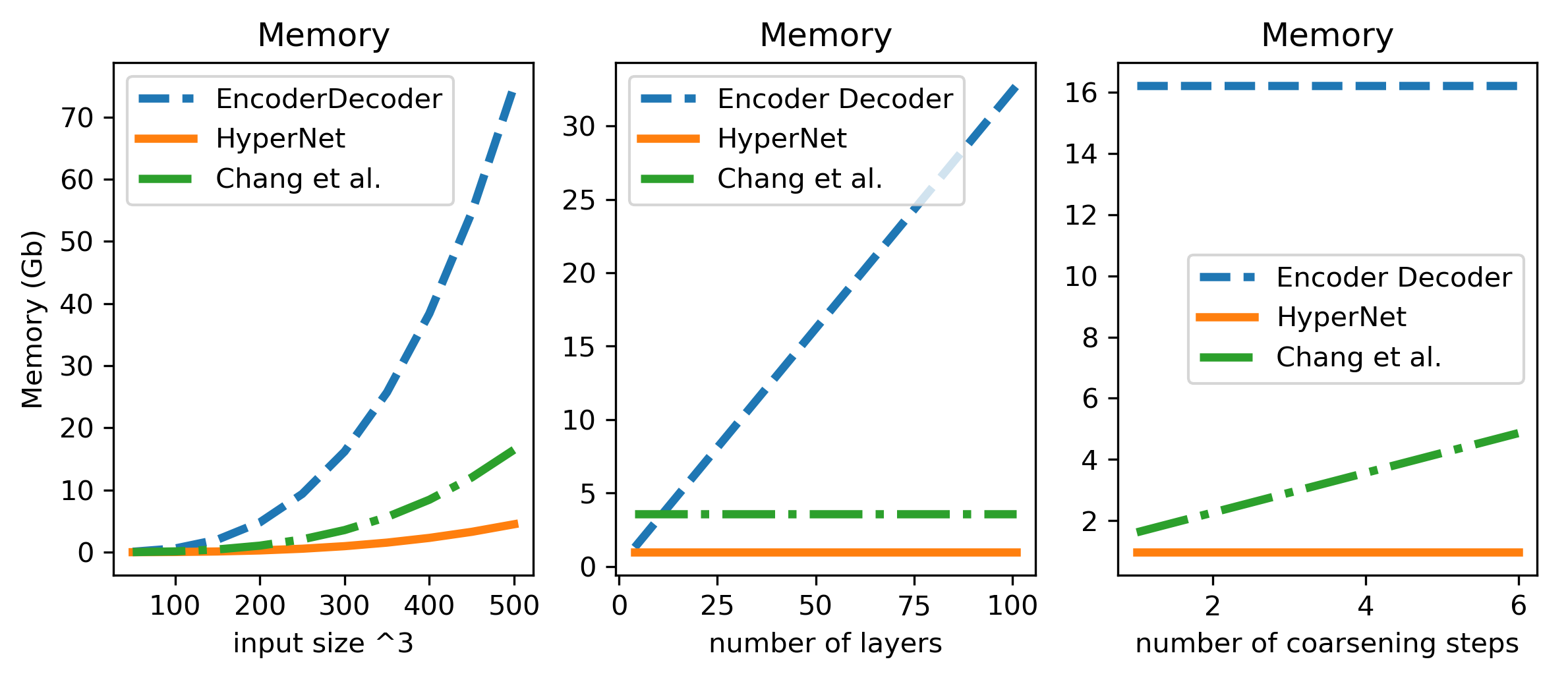}
        \caption{Memory comparison related to the activations (network states) stored in the forward-pass through the network for computing gradients. Figure displays a
Resnet-based encoder-decoder network, block-reversible network \protect\cite{Chang2017Reversible}, and the proposed fully reversible network. Left: assumes $50$ layer network with $4$ coarsening and refinement stages. Middle: $300^3$ input and fixed number of coarsening steps. Right: $50$ layers and $300^3$ input size.}
        \label{fig:NetworkMemory3D}
\end{figure}

Various researchers developed neural networks based on partial differential equations (PDEs) to gain insight and control of the stability with an increasing number of layers, in order to avoid exploding and vanishing gradients. By basing the networks on certain PDEs, one can achieve reversibility, such that not all network states need to be stored to be able to compute a gradient. Published networks in \cite{Chang2017Reversible,GomezEtAl2017} are block-reversible, meaning that the network is reversible only in between pooling operations that change resolution and the number of channels. This means that block reversible networks still require storing the network state before coarsening/refining layers. We propose a fully reversible network where the coarsening/refining operations are also reversible/invertible. The required storage to compute gradients becomes independent of the network depth, as well as the number of coarsening/refining operations. This allows us to work with arbitrarily large images as there are no memory limitations preventing the use of a network that achieves a complete receptive field, as opposed to a block reversible network where memory limitations restrict the possible number of coarsening/refining layers as demonstrated in Figure \ref{fig:NetworkMemory3D}.

In this work we build off of \cite{Chang2017Reversible} who propose reversible architectures motivated by the leap frog discretization of second order hyperbolic dynamical systems. 
This work benefits from known stability criteria, given that it is based off of well understood numerical methods with known regions of stability.
NeuralODE \cite{neuroODE} similarly use dynamical systems to construct reversible architectures, however they propose integrating a first order system forward in time, and then integrating a different first order system backward in time to recover the input. This has been shown to lead to numerical instabilities upon discretization \cite{gholami2019anode}.

\cite{GomezEtAl2017} construct RevNet, which is an algebraiclly reversible network by combing sequences of invertible layers, that is similar to the Hamiltonian networks described in \cite{Chang2017Reversible}. 
These algebraiclly reversible networks are extending in \cite{irevnet} using a form of invertible pooling called \textit{pixel-shuffle} to form fully reversible classifiers.
While these architectures are algebraically reversible, they do not have explicit stability criteria and are susceptible to growing error from numerical instabilities.
In \cite{irevnet} the stability of the network is empirically evaluated for the given task by comparing the reconstruction of the input, however this observation is specific to the given task and network architecture, and provides no guarantees of stability in other settings.

We propose to use the discrete wavelet transform as an invertible pooling operator to coarsen the image and increase the number of channels, as well as the reverse operation. 
While the wavelet transform has been used before in neural networks as a fixed operator \cite{WaveletNetwork1,fujieda2017wavelet}, this work is the first to propose their use to achieve full reversibility.
RevNet has again been extended in the new concurrent work \cite{iUnet}, which addresses the checker board artifacts from pixel shuffle by constructing a fully reversible iUNet. 
This work similarly uses a learnable form of invertible pooling based off of the discrete wavelet transform.
Simple invertible pooling operations have previously been proposed \cite{DinhSB16,irevnet}, however their motivation is not based on PDEs so they do not provide the same insights on network stability and they were not proposed in the context of reducing the memory footprint.

Finally, we show how a reversible dynamics can be applied in the context of Variational Auto Encoder (VAE), in a manor that is a kin to the truncated SVD. This new application of reversible networks allows us to control the amount of information dropped out in the VAE which is one of the main difficulties when training a VAE.

 \section{Building Blocks of the Fully Hyperbolic Network}
\label{buildingBlocks}

\subsection{Reversible Architectures}
We start by reviewing the architecture of a canonical ResNet as presented in \cite{he2016deep}.
A ResNet is a particular architecture of a neural network. Let $\bfY_0$ represent
the data or initial state, which we view as a matrix where each column (vector) in the matrix is a particular datum. Let us focus on one datum. For problems in imaging we can reorganize the datum as a tensor that represents a vector image with $n_c$ channels and a resolution of $n_x \times n_y$.

The initial state $\bfY_0$ is transformed by the network using the following
expression
\begin{eqnarray}
\label{resnet}
\bfY_{j+1} = \bfW_j \bfY_j +  f( \bfY_j,\bftheta_j)
\end{eqnarray}
Here, $\bfY_j$ is the state at layer $j$ and $\bftheta_j$ are network parameters, e.g. convolution kernels, normalization parameters, and biases, and finally $f$ is a nonlinear function. The matrix $\bfW_j$ is the identity matrix if the number of channels in $\bfY_{j+1}$ is equal to the number of channels in $\bfY_j$ and is chosen as zero otherwise.
In this work we particularly use convolutional neural networks with $\bftheta_j = \{\bfK_j,\bfb_j\}$, where $\bfK_j$ are convolution kernels and $\bfb_j$ are biases (for batch norm we do not use bias but use batch norm parameters). We choose a symmetric nonlinear function that can be expressed as
$$  f( \bfY_j,\bftheta_j) = -\bfK_j^{\top} f(\bfK_j \bfY_j + \bfb_j), $$
which has been shown to have favorable theoretical and experimental properties \cite{RuthottoHaber2018}.

For many, if not most, applications the number of channels grows with depth, while at the same time the state becomes coarser. Since this is typically done using non-invertible coarsening or pooling the network loses information. Consider average pooling, which corresponds to linear interpolation and has been proven to work well on a number of applications. Despite its success, average or max pooling is not reversible, meaning that given the state $\bfY_j$ we can compute $\bfY_{j+1}$, however if we are given $\bfY_{j+1}$ it is generally impossible to compute $\bfY_j$.

 A reversible network has a few advantages. One of the most important being that it does not require the storage of the activations when computing the gradients, allowing for arbitrary long networks independent of memory \cite{Chang2017Reversible,GomezEtAl2017}. In order to obtain a reversible network it was proposed in \cite{Chang2017Reversible} to use a hyperbolic network with the form
\begin{eqnarray}
\label{resnethyper}
\bfY_{j+1} = 2 \bfW_j\bfY_j -  \bfW_{j-1} \bfY_{j-1} +  f( \bfW_j \bfY_j,\bftheta_j)
\end{eqnarray}
Here, again, we use $\bfW_{j-1}$ and $\bfW_j$ when the resolution is changed and the number of
 channels are increased and set $\bfW_{j-1} = \bfW_j = \bfI$ when the resolution does not change.
 The network is clearly reversible since, as long as the number of channels are not changed and $\bfW=\bfI$, given $\bfY_{j+1}$ and $\bfY_j$ it is straight forward to compute $\bfY_{j-1}$. Reversibility allows for the computation of the gradient {\em without} the storage of the activations. This is done by stepping backwards and computing the states and their derivatives in the backward pass. This of course does not come for free as the computational cost for computing derivatives is doubled. However, for problems with deep networks, memory is typically the limiting factor in training and not the computational cost.
 
 The origin of such a network can be traced to a nonlinear Telegraph equation \cite{Zhou2018Telegraph} which can be written as
 \begin{eqnarray}
 \label{telegraph}
 \ddot{\bfY} = f(\bfY,\bftheta(t)).
 \end{eqnarray}
 The Telegraph equation describes the propagation of signals over physical and biological networks and therefore it is straight forward to extend its use to deep neural networks where signals are propagated over artificial networks. 
 A leapfrog finite difference discretization of the second derivative reads
 $$  \ddot{\bfY}  \approx {\frac 1 {h^2}} \left(\bfY_{j+1} - 2 \bfY_j + \bfY_{j-1} \right). $$
This leads to the proposed network \eqref{resnethyper} where the $h^2$ term is absorbed into the network parameters. The network is named hyperbolic as it imitates a nonlinear hyperbolic differential equation \eqref{telegraph}. 
 
  For hyperbolic networks where $\bfW_j$ is not invertible for every layer, the network is only reversible in blocks where $\bfW_{j} = \bfW_{j-1} = \bfI$. In such cases, one can compute $\bfY_{j-1}$ given $\bfY_{j+1}$ and $\bfY_j$, removing the need to store activations as they can recomputed in the backward pass. However, since activations still need to be stored in cases where $\bfW_j$ is not invertible the memory footprint depends on the number of non invertible $\bfW_j$ , as shown in Figure \ref{fig:NetworkMemory3D}. In the context of a CNN with non-invertible $\bfW_j$, e.g. average pooling, this means that one would need to store the two states preceeding every layer that changes the resolution of the state. In some cases, such as encoder-decoder architectures, this can lead to a significant increase in the network's memory footprint, see Figure \ref{fig:NetworkMemory3D}. For large scale problems and deep networks, where the memory required to store activations exceeds what is possible with current hardware, this may be the only feasible way to compute gradients. 

\begin{figure}
\begin{center}
\includegraphics[width=8cm]{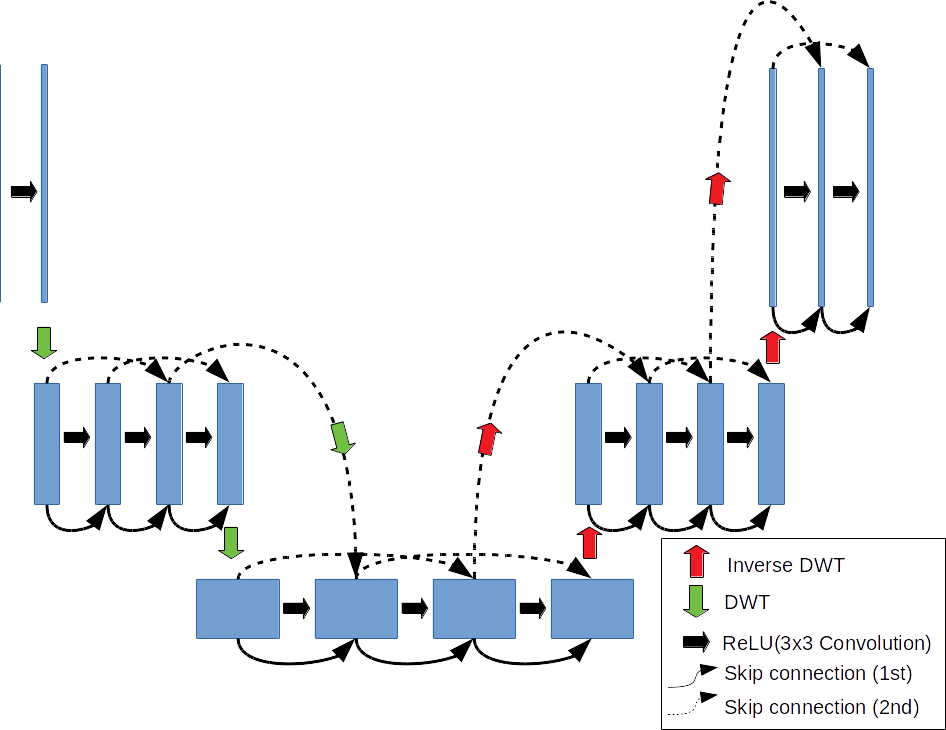}
\caption{A sketch of a 12 layer hyperbolic network for high dimensional output problems. The first layer opens the image up to the desired size of the output. The hyperbolic network uses two skip connections to compute the next layer. The DWT and its inverse are used to coarsen the state and increase the number of channels without losing information. In cases where the skip connection connects states that are different dimensions, the appropriate transform is applied. 
\label{network_diagram}}
\end{center}
\end{figure}


\subsection{Reversible Coarsening and Refinement}
Let us now introduce a hyperbolic network that is fully reversible and overcomes the above issues. We are interested in problems where both the input and the output are large and dense, e.g. image and video segmentation, as this is where the memory footprint of the network is particularly important. For these problems it is common to use an encoder-decoder architecture \cite{Shelhamer2017FCN}. When such a network is applied to the data, going from the low resolution latent state to high resolution output requires interpolation. However, interpolating the low level features introduces interpolation artifacts and damps high frequencies. To regain high frequencies, the interpolated image is then supplemented with previous states, such as at different scales \cite{PSPNetZhao2017} or higher resolutions \cite{UNET2015}. Unfortunately, in order to have a reversible encoder-decoder network, it is necessary to remove irregular skip connections that introduce dependencies between distant states of the network. The removal of these skip connections and the introduction of reversible pooling is required in order to design a reversible network with constant memory requirements during training, independent of network depth. Additionally, the latent space of a hyperbolic network will not have lost any information, so it will not be necessary to use previous states to re-introduce information via long skip connections.

To this end we focus our attention on the layers where the resolution and number of channels are changed.
Let $\widehat \bfY_j$ be the $j$-th state obtained at layer $j$. Our goal is to obtain a new state $\bfY_j$ that has a coarser resolution with more channels. This of course can be done by simply convolving $\widehat \bfY_j$ by a few kernels and then, coarsening each one of them which can be written as
\begin{eqnarray}
\label{bottleneck}
\bfY_j = \begin{pmatrix}   \bfP &           &       &  \\ 
                                                 & \bfP    &      &    \\
                                                 &            & \ddots  &   \\
                                                 &            &     & \bfP
                                                 \end{pmatrix} 
\begin{pmatrix} \bfA_1 \\ \bfA_2 \\ \vdots \\ \bfA_n \end{pmatrix} \widehat \bfY_j
\end{eqnarray}
Here, $\bfA_i$ are convolution matrices and $\bfP$ are restriction matrices.
The resulting state $\bfY_j$ has now $n$ channels, and each channel has a lower resolution compared to the original state.

Consider the matrix 
\begin{eqnarray}
\label{bottleneck2}
\bfW_j =  
\begin{pmatrix} \bfP \bfA_1 \\ \bfP \bfA_2 \\ \vdots \\ \bfP \bfA_n \end{pmatrix} 
\end{eqnarray}
In general, the matrix, $\bfW_j$ is rectangular or not invertible. However, if we construct the
matrix in a way that it is square and invertible then it is possible to decrease the resolution and add channels or to {\em increase the resolution} and reduce the number of channels without losing any information and without interpolation.
This will enable us to compute $\bfY_{j-1}$ given $\bfY_{j+1}$ and $\bfY_j$ even when $\bfW_{j} \neq \bfW_{j-1} \neq \bfI$, meaning that we will have a fully reversible network.

While it is possible to learn the matrices $\bfA_i$, this may add considerable complexity to the method. This is because, while it is straight forward to build $\bfW_j$ as square, it is not obvious how to enforce its invertibility. Furthermore, even if the matrix is invertible, inverting it may not be simple, making the process too expensive for practical purposes. Although it is difficult to learn an appropriate matrix $\bfW$, it is possible to choose one that posses all the above qualities.

We propose to use the discrete wavelet transformation (DWT) as the invertible operator $\bfW$ that coarsens and increases the number of channels.
The DWT is a linear transformation of a discrete grid function \cite{WaveletReview}, and is commonly used in image processing. In its simplest form at a single level, the DWT applies four filters to an image and decomposes it into four coarser images each containing distinct information. The important point is that the DWT is invertible, that is, it is possible to use the four low resolution images in order to explicitly and exactly reconstruct the fine resolution image. 

While there are many possible wavelets that can be used, here we chose first to use the Haar wavelet. The Haar wavelet uses four simple filters and can be interpreted as simple convolution with subsampling (stride). The first is simply an averaging kernel that performs image sub-sampling. The final three are a vertical derivative, a horizontal derivative, and a diagonal derivative.
In essence, the wavelet transform gives a recipe to either coarsen the image and increase the number of channels or reduce the number of channels and increase the resolution. This is exactly the property that is needed for our network.

Using the above ingredients the hyperbolic network is simply \eqref{resnethyper} with $\bfW_j$ being the wavelet transform. 
For the dense input and output problems that we are considering, we require a downward pass where the image is coarsened and the number of channels are increased and an upward pass where the image is refined and the number of channels are decreased as follows
\begin{align}
\nonumber \bfY_{j+1} &= 2 \bfW_j\bfY_j -  \bfW_{j-1} \bfY_{j-1} +  f( \bfW_i \bfY_j,\bftheta_j)
\\  &\quad j=1\ldots n \\
\nonumber \bfY_{j+1} &= 2 \bfW_j^{-1}\bfY_j -  \bfW_{j-1}^{-1} \bfY_{j-1} +  f( \bfW_i^{-1} \bfY_j,\bftheta_j)\\ 
&\quad j=n+1\ldots 2n.
\label{eqn:hypernet}
\end{align}

The final output has the same resolution as the input image and, since both parts of the network are reversible, the entire network is reversible.

\subsection{Generalized Learnable Wavelets}
Selecting $\bfW$ to be a specific wavelet, such as the previously mentioned Haar wavelet, simplifies things by ensuring that we have a known inverse due to the orthoganality of the matrix, however the calculation of these specific coefficients comes from additional constraints that may not always be applicable to the dataset we are training or to that particular layer. 
For the class of discrete wavelets discovered by Daubechies, these values are determined for specific filter lengths by solving systems of equations that enforce orthogonality and a zero response to a particular order of smoothness in the signal known as approximation condition \cite{dbOrthonormal}.
The Daubechies wavelet with four coefficients, $\{c_0, c_1, c_2, c_3\}$, is found by solving the following system of four equations where we constrain for orthogonality, and constrain for an approximation condition of $p=2$ \cite{ptvf}.

\begin{subequations}
\label{eq:wvtcond}
\begin{eqnarray}
c_0^2 + c_1^2 + c_2^2 + c_3^2 = 1, \quad c_2c_0 + c_3c_1 = 0 \label{eq:ortho} \\
c_3 - c_2 + c_1 - c_0 = 0, \quad  0c_3 - 1c_2 + 2c_1 - 3c_0 = 0 \label{eq:order}
\end{eqnarray}
\end{subequations}

Seeing as \ref{eq:order} are chosen based on an assumption about the smoothness of the signal, and only \ref{eq:ortho} are required for invertibility, we propose learning wavelets by solving \ref{eq:ortho} and using the two remaining degrees of freedom to learn the best wavelet for each particular layer of the network. 
This gives us extra flexibility to use a wavelet that fits the data at hand.

%
%

\section{Reversible Networks for Variational Auto Encoders}
\label{AE}

In this section we describe an approach for using the proposed fully hyperbolic networks to construct a Variational Auto Encoder (VAE).

\begin{figure}[h]
\begin{center}
\begin{tikzpicture}[scale=1.0]

 \pgfsetfillopacity{1.0}
 \fill[color=mycolor] (10,-3) circle (8mm);
 \fill[color=mycolor]  plot[smooth cycle] coordinates{(9,0) (10.5,1.5) (10.2,1.8)  (10.7,2.3)  (12,2)  (11.5,1.5)  (11.0,-1.3)};
  \fill[color=mycolor]  plot[smooth cycle] coordinates{(2,0) (3.5,1.5) (3.2,1.8)  (3.7,2.3)  (5,2)  (4.5,1.5)  (4.0,-1.3)};
  
 \draw [->](4.9,-3) -- (9,0);
 \draw (8.8,-1.5) node {$ \widehat\bfX = D(\bfZ,\bftheta)$ };

 \draw [densely dotted](5,2) -- (10.3, 2);
 \draw (7.6,1.5) node {$\|\bfX - \widehat \bfX\|^2$ };

 \draw [densely dotted](4.9,-3) -- (9.3, -3);
 \draw (7,-3.5) node {${\mathbb{ KL}}(\bfZ, \bfT)$ };

\fill[color=mycolor]  plot[smooth cycle] coordinates{ (2.7,-3.7)  (4.2,-2.2)  (5,-3.4)  (4.0,-3.9)  (4.0,-3.6)};

\draw (10,-3) node {${\cal T}$ };
 \draw (4,-3) node {$\cal Z$ };
 \draw (10.5,1) node {$\widehat{\cal X}$ };
 \draw (3.5,1) node {${\cal X}$ };

 \draw [<-](4.2,-2.2) -- (4.0,-1.3);
 \draw (2.5,-1.9) node {$\bfZ = E(\bfX,\bfzeta)$ };
 
\end{tikzpicture}

\caption{A diagram showing the typical structure of a variational auto-encoder. The encoder $E(\cdot,\bfzeta)$ is used to transform input data $\bfX \in \cal X$ to the latent space $\bfZ \in \cal Z$. The data in the latent space is compared with an optimal distribution ${\cal T}$. The decoder $D(\cdot,\bftheta)$ maps from the data from the latent space to an approximation of the input $\widehat \bfX \in \cal X$. The loss is determined by the penalizing the reconstruction, shown here with the $\ell_2$ norm, and penalizing the distance between the latent space $\cal Z$ and the target space $ \cal T$.}
\label{vae}
\end{center}
\end{figure}
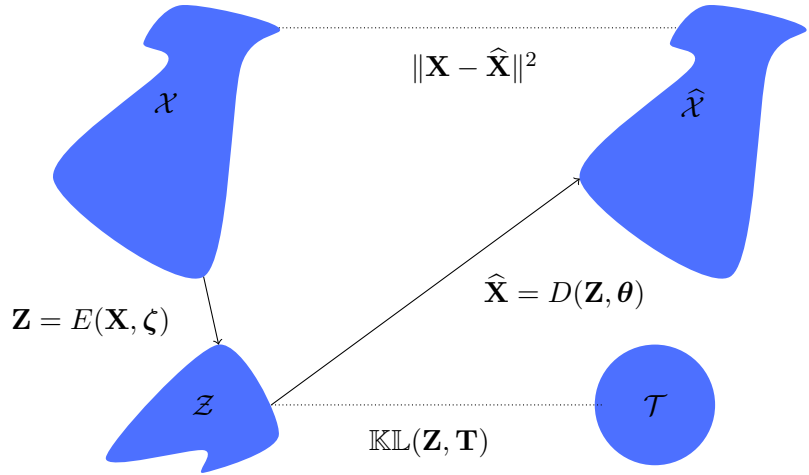

VAE's can be seen as generative models.
Their aims to use existing data to generate new data that is "similar" to the given data. 
Typically the data is modelled as points drawn from a complex probability distribution in very high dimensions and the goal of a generative model is to be able to sample from such a probability density function (see \cite{SalimansGZCRC16,karras2017progressive,brock2018large,karras2018stylebased} and references within).
VAE's \cite{kingma2013autoencoding} are a class of generative models that are based on Auto-Encoders (AE). 
Here, the idea is to use an AE in order to train a transformation (an Encoder) from the complex input space, ${\cal X}$, to a latent space, $\widehat {\cal Z}$, and then train a second transformation (a Decoder) from the latent space, $\widehat {\cal Z}$, back to a complex space, $\widehat {\cal X}$, such that it approximates the input space ${\cal X}$. 
Typically the encoder and decoder take the form of separate neural networks that reduce the dimensionality of the network state, such that the latent space is smaller than the input space, effectively learning a compressed latent state. 
The latent space, $\widehat {\cal Z}$, in an AE is not simple, that is, it can be highly non-convex.
The variational part of the VAE is obtained by penalizing latent spaces such that they approximate a simple space, typically a Gaussian space, ${\cal T}$. 
A sketch of a VAE is given in Figure~\ref{vae} where the Kullback–Leibler (KL) Divergence is used to compute the distance between the latent space and the ideal space.  

\bigskip

Here 
we propose to use a forward reversible network as the encoder, in order to get to a latent space and the subsequent backward time integration from the latent space to the image as the decoder.

The encoding and decoding steps are non-lossy and therefore, if we use them as is, no dimensionality reduction can be obtained. In order to reduce the dimension of the network use use a projection on the latent space that reduces the amount of information in the space. This can be written as follows 
\begin{subequations}
 \label{encodecode}
 \begin{eqnarray}
 \label{encode}
 {\rm Encoder:} && \widehat \bfZ = E(\bfX,\bftheta,\bfP) =  \bfP^{\top} F(\bfX,\bftheta) \\
  \label{decode}
 {\rm Decoder:} && \widehat \bfX =  E^{\dag}(\widehat \bfZ,\bftheta,\bfP) =   F^{-1}(\bfP \widehat\bfZ,\bftheta) 
 \end{eqnarray}
 \end{subequations}
Here $F$ is a reversible network, $F^{-1}$ is its inverse and $\bfP^{\top}$ is a learnable projection matrix that reduce the dimension of the problem.

The process proposed above is a nonlinear analog to the truncated SVD. Here, the encoder is equivalent to the projection by an orthogonal transformation. Similarly to the SVD it retains information. The multiplication by the projection matrix $\bfP$ is truncating the space and the encoder is an inverse of the decoder.

Note that if we choose $\bfP = \bfI$ the AE is exact, that is, there is no loss of information. In order to reduce the dimensionality we proposed to minimize the following objective function

\begin{eqnarray}
\label{eq:objVAE}
  	\min_{\bfP,\bftheta} \quad \| F^{-1}(\bfP  \bfP^{\top}F(\bfX,\bftheta),\bftheta) - \bfX\|^2
  	+ \alpha \|\bfP\|_{\cal C} + \beta \varepsilon(\bfP^{\top}F(\bfX,\bftheta)) \\
  	\nonumber
\end{eqnarray}
Here, $\| \cdot \|_{\cal C}$ is a 1-2 group norm. It is used in order to obtain a matrix $\bfP$ with very few non-zero rows. The last term is a penalty term that measures the Gaussianity of the latent space. This can be done by the KL divergence as proposed in \cite{kingma2013autoencoding}, however, we have found experimentally that a better way to steer the learning process towards a Gaussian distribution is to use the function proposed in \cite{Szekely_testingfor}.

\section{Optimization}
\label{sec:opt}
Training a neural network implies  obtaining network parameters, $\bftheta$, (weights) that minimize the mismatch between the output of the neural network at the last layer ($n$), $\bfY_n$, and the labels $\bfC$, which is done by minimizing a desired loss function, $\ell(\bfC,\bfY_n)$, over the network parameters.

Reverse-mode automatic differentiation is the most commonly used technique to compute a gradient of a function that includes a neural network. Reverse-mode automatic differentiation is an implementation of the backpropagation algorithm, or adjoint-state method. While practically convenient, automatic differentiation stores the network states for every channel at every layer in the network. This leads to prohibitive memory requirements when the input is large, e.g. 4D data such as RGB video or time-lapse hyperspectral data, particularly when working on relatively low-memory GPUs. To take advantage of the fully reversible structure of the hyperbolic network design with invertible pooling, we need a different implementation of the backpropagation algorithm. 

Note that the reversible network can be written as going backwards in time
$$ \bfY_{n-1} = 2  \bfW_{n-1}^{-1} \left(\bfW_n \bfY_n - \bfY_{n+1} + h^2f(\bfW_n\bfY_{n},\bftheta_n) \right) $$

This implies that by doubling the computation and computing $\bfY_j, j=n-2, \ldots, 2$ backwards, we can recompute the activations without storing them. 
An implementation of the idea is presented in 
Algorithm \ref{grad_low_mem}.
\begin{algorithm}[hbt]
\begin{algorithmic}
 \STATE{$\bfY_{1} = \bfD$, $\bfY_2 = \bfD$ //initialization}
 \STATE{//forward propagation}
 \FOR{$j=3,\cdots,n$}
  \STATE{$\bfY_{j+1} = 2 \bfW_j\bfY_j -  \bfW_{j-1} \bfY_{j-1} +  f( \bfW_j \bfY_j,\bftheta_j) $}
 \ENDFOR
\STATE{save $\bfY_{n}$ and $\bfY_{n-1}$}
\STATE{$\bfQ_{n} = \grad_{\bfY_n}l(\bfC,\bfY_n)$}
\STATE{$\bfQ_{n-1} = 2 \bfQ^\top_n \big[ \bfW_{n-1} + \grad_{\bfY_{n-1}}f( \bfW_{n-1} \bfY_{n-1},\bftheta_{n-1})\textbf]$}
\STATE{//backward propagation}
 \FOR{$j=n-1,\cdots,3$}
 \STATE{$\grad_{\bftheta_{j}} L =  \bfQ_{j+1}^\top \grad_{\bftheta_{j}} f( \bfW_j \bfY_j,\bftheta_j)$}
\STATE{$\bfY_{j-1} = \bfW_{j-1}^{-1} \bigg[ 2 \bfW_{j} \bfY_{j} + f( \bfW_j \bfY_j,\bftheta_j) - \bfY_{j+1} \bigg]$}
 \STATE{$ \bfQ^\top_{j-1} = 2 \bfQ^\top_{j} \big[ \bfW_{j-1} +  \grad_{\bfY_{j-1}}f( \bfW_{j-1} \bfY_{j-1},\bftheta_{j-1}) \big] + \bfQ^\top_{j+1} \bfW_{j-1}$}
\STATE{$\bfQ_{j+1} \leftarrow \bfQ_{j}$, \: $\bfQ_{j} \leftarrow \bfQ_{j-1}$, \: $\bfY_{j+1} \leftarrow \bfY_{j}$, \: $\bfY_{j} \leftarrow \bfY_{j-1}$}
\ENDFOR
\RETURN{gradient of network w.r.t kernels }
 \caption{Low-memory implementation of gradient computations by re-computing the fields on-the-fly via network reversibility.}
\label{grad_low_mem}
\end{algorithmic}
\end{algorithm}
Here we use the matrices $\bfQ_j$ as the states of the propagated backward network that are computed on the fly. They require the evaluation of the Jacobian vector product of the activations over a {\em single} time step and this can be done either analytically or by automatic differentiation. 

\section{Numerical Experiments with Hyperbolic Networks}
\label{experiments}

In this section we experiment with our architecture for problems that range from high input and video segmentation to VAE's.

\subsection{Memory Restricted Applications}
We experiment with our network on two different problems that require large input and output. The first is the estimation of land-use change from 4D time-lapse hyperspectral data and the second segmentation of RGB video. We show in these numerical experiments that the DWT is an appropriate choice for an invertible pooling operator. The low and constant memory requirements of our fully hyperbolic network enables training on a single GPU, whereas memory requirements for equivalent non-reversible networks would be problematic.

\begin{table}[htb]
\begin{center}
\begin{tabular}{|l|cc|}
\hline
Memory in MB. & Fully Reversible & Non-Reversible   \\
\hline\hline
Hyperspectral & 798 & 5057 \\
Video & 528   & 3693  \\
\hline
\end{tabular}
\end{center}
\caption{The memory requirements for the states for both numerical examples.}
\label{tab:NetworkMem}
\end{table}

The scope of these examples is different from most semantic segmentation tasks. Commonly, networks train on large numbers of examples with full supervision (e.g., fully annotated videos) and provide quick inference afterward on a new example. Here, we show that we can train a network on a single example, given a partial labeling. We never have access to any fully annotated examples. The task is thus a type of interpolation and extrapolation of the provided labels, guided by the data. We note that the benefit of not requiring a training dataset comes at the cost of not having a real-time segmentation method. Using partial annotations to complete the 3D annotation is also referred to as semi-automated or semi-supervised in some literature. The task of completing partially annotated 3D data volumes also appears in medical image segmentation \cite{Unet3D} and geophysical data processing \cite{doi:10.1190/INT-2018-0225.1}.


\subsubsection{Time-lapse Hyperspectral Segmentation}

Hyperspectral data is intrinsically 3D: two spatial and one frequency dimensions. This becomes 4D when we collect data at multiple times. Because the number of frequencies easily exceeds $100$, just as the number of pixels in each spatial dimension, this is a large-scale problem in terms of memory. Because of memory limitations primarily, 3D convolutional neural networks for hyperspectral segmentation typically work with either, pseudo-3D convolutions, lossy dimensionality reduced data as input to the network, networks that map a data patch to a classification of the central pixel, data inputs not exceeding $27 \times 27 \times 200$, and often less than five layers, see e.g., \cite{doi:10.1080/2150704X.2019.1686780,rs12010188,rs9010067}.

\begin{figure}[htb]
\begin{center}
    	\begin{subfigure}[b]{0.3\linewidth}
 		\includegraphics[width=\textwidth]{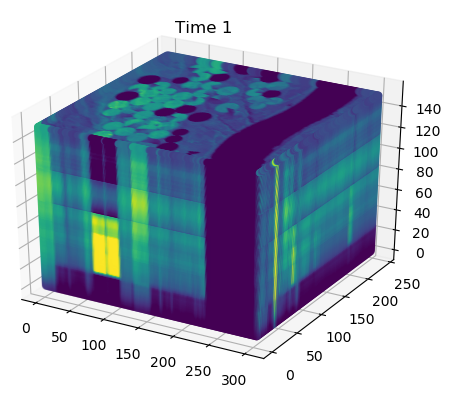}
 		\caption{}
 		\label{fig:Figure1a}
 	\end{subfigure}
 	\begin{subfigure}[b]{0.3\linewidth}
 		\includegraphics[width=\textwidth]{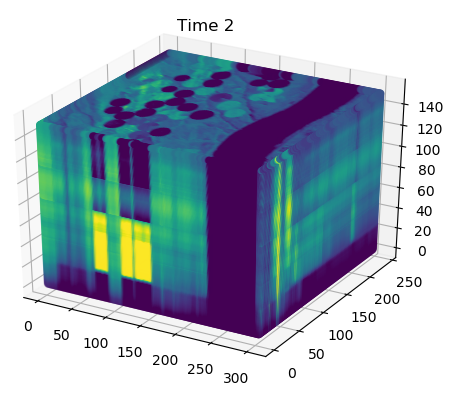}
 		\caption{}
 		\label{fig:Figure1b}
 	\end{subfigure}
 	   \begin{subfigure}[b]{0.6\linewidth}
       \includegraphics[width=\textwidth]{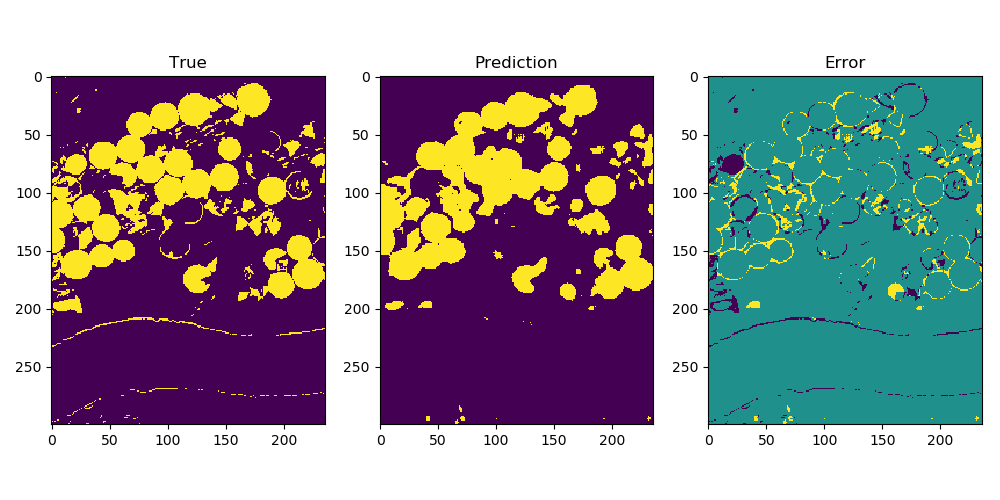}
       \caption{}
        \label{fig:Figure1c}
   \end{subfigure}
          \caption{(a) and (b) hyperspectral data collected at two different times. (c) True time-lapse change, final thresholded prediction, and the error. The main source of error is at the boundaries for the center-pivots, only two circles are classified incorrectly.}
   \label{fig:prediction_threshold}
   \end{center}
\end{figure}

Using this example we highlight that \emph{a}) fully reversible networks can take the whole 4D time-lapse hyperspectral data volume as input to the network; and \emph{b}) while the input size for a fully reversible network is the same as the input size, we can still use this network type to learn mappings from 4D input to 2D output. Figure \ref{fig:prediction_threshold} shows the hyperspectral data \cite{doi:10.1080/01431161.2018.1466079}, collected at two different times. We use the full $304 \times 240 \times 152 \times 2$ data tensor as input to our network. 

The data in Figure \ref{fig:Figure1a}, \ref{fig:Figure1b}) is the entire dataset and there is no other training data. The goal is to classify each pixel in a plan-view of the data based on some of the ground truth. About $8.8 \%$ of the pixels are labeled for training, which is significantly less than most research on hyperspectral imaging for agricultural classification that often use between $30\% - 50\%$ of the labeled pixels for training \cite{C7729859,M8297014,doi:10.1080/2150704X.2019.1681598,rs9010067,rs12010188}. This ground truth could be obtained by manual data analysis by an expert, or by a surface inspection in the field.

 
Our network learns a map from $\mathbb{R}^{n_x \times n_y \times n_\text{freq} \times n_\text{time}}$ to 4D output, but we measure the loss at a single 2D $x-y$ slice only. In other words, the 4D label tensor has known entries at a single slice. While it is common to to treat hyperspectral data as a 2D image with many channels, in our time-lapse context we deal with the frequencies as the third dimension in the data volume, and use the different channels for the data collected at different times. We increase the number of channels of the input while decreasing the size and preserving the information via two 3D Haar transforms, see Table \ref{network_design} for the details of the network. Taking the transforms beforehand saves two transforms and shortens the network a bit, saving computational time at every iteration.

In Table \ref{tab:NetworkMem}, we list the memory that is required for the network states $\bfY_i$ in order to compute a gradient for each example. These memory requirements will not increase for our fully reversible network is we add more layers to the network, as opposed to non-reversible networks that will exhibit a linear memory growth in that case. The final thresholded prediction, true mask, and errors are displayed in Figure \ref{fig:prediction_threshold}. The results are mainly correct, except one of the center-pivot circles is classified incorrectly, as well as some boundary artifacts. Our network achieves a mean IOU of $82.5$ on this binary task, with a positive IOU of $73.8 $ and and negative IOU of $ 91.1$.

\begin{table}[htb]
\centering
\begin{tabular}{|r|r|c|}
\hline
\textbf{Hyperspectral: }& & \\
Layer \# &  Feature size &  \# of channels\\
\hline
1-5   & $76 \times 60 \times 38$ & 384\\
6-11   & $152 \times 120 \times 76$ & 48  \\	
12-19   & $304 \times 240 \times 152$ & 6 \\	
\hline
\hline
\textbf{Video:} & &\\
Layer \# &  Feature size &  \# of channels\\
\hline
1-5   & $60 \times 106 \times 18$ & 384 \\
6-11   & $120 \times 212 \times 36$ & 48 \\	
12-19   & $240 \times 424 \times 72$ & 6 \\	
\hline
\end{tabular}
\caption{Details of the fully reversible hyperbolic network used for both the hyperspectral and video example. All convolutional kernels are of size $3\times3 \times 3$. We apply two 3D Haar transforms to the data to generate data with more channels, but smaller in terms of the number of spatial/temporal/frequency elements.}\label{network_design}
\end{table}


\begin{figure*}[]
\begin{center}
    \begin{subfigure}[t]{0.32\textwidth}
        \includegraphics[width=\textwidth]{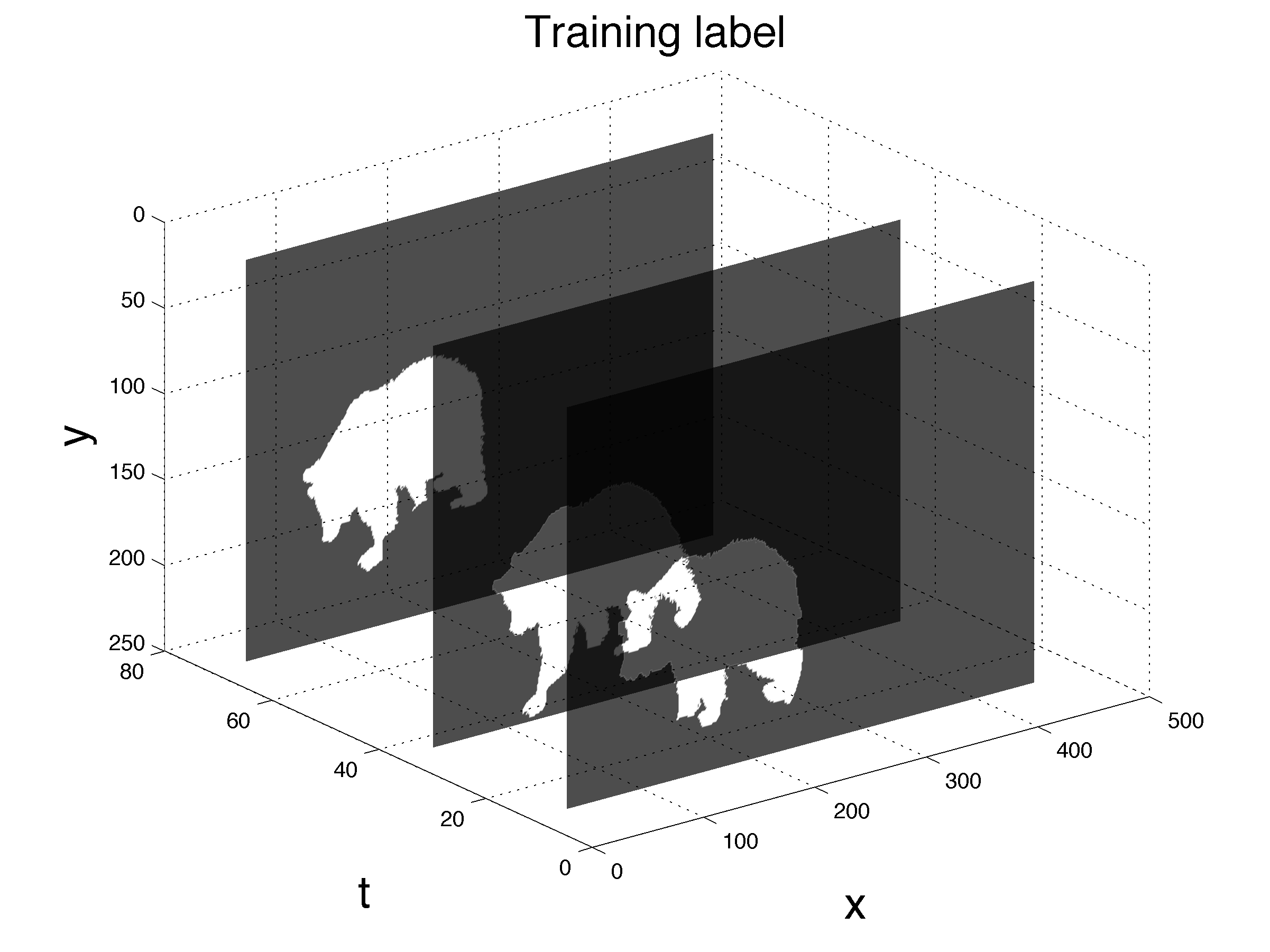}
        \label{fig:Bear_Label3D}    
    \end{subfigure} 
    \begin{subfigure}[t]{0.32\textwidth}
        \includegraphics[width=\textwidth]{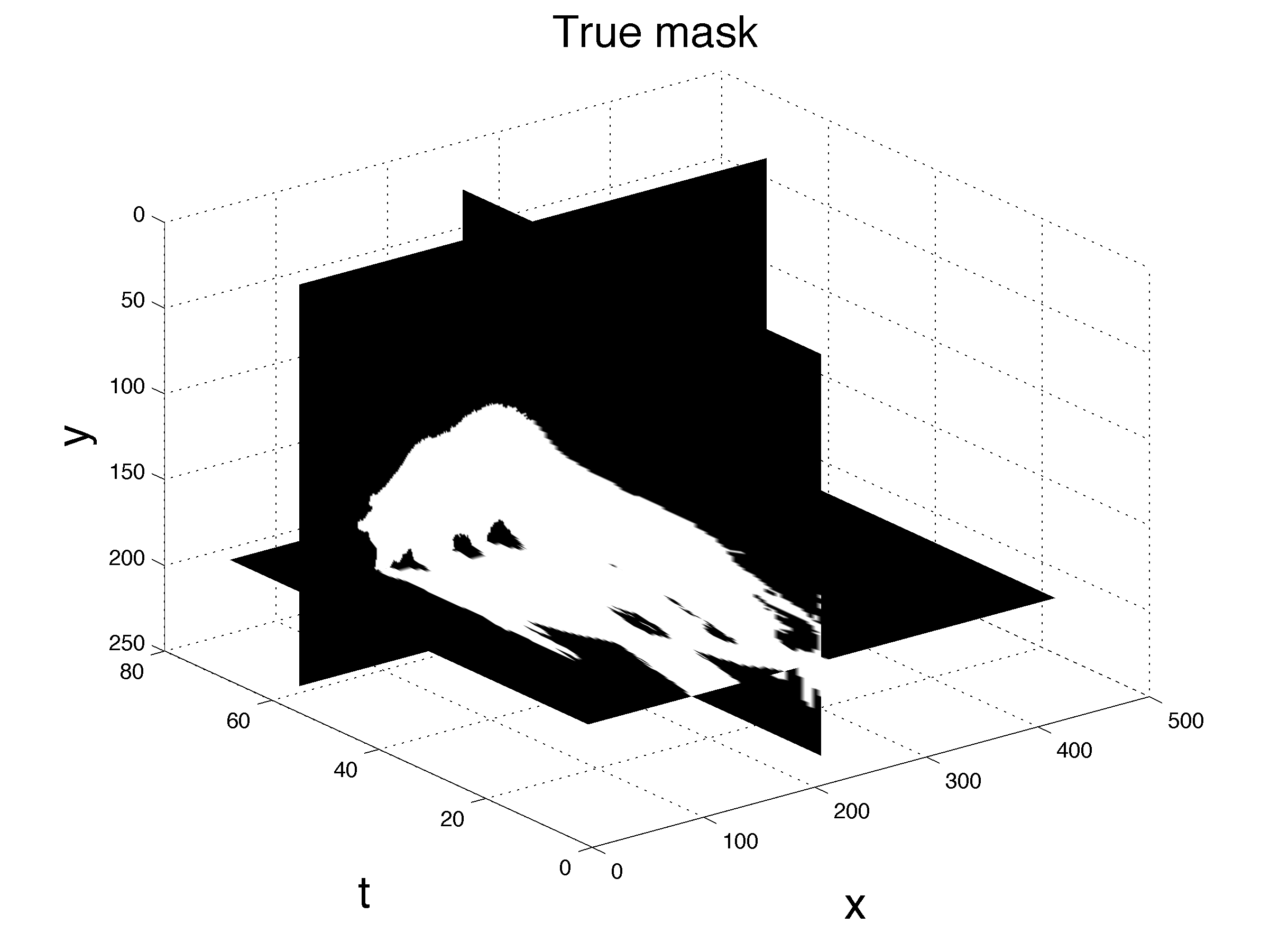}
        \label{fig:true_mask}
    \end{subfigure}
    \begin{subfigure}[t]{0.32\textwidth}
        \includegraphics[width=\textwidth]{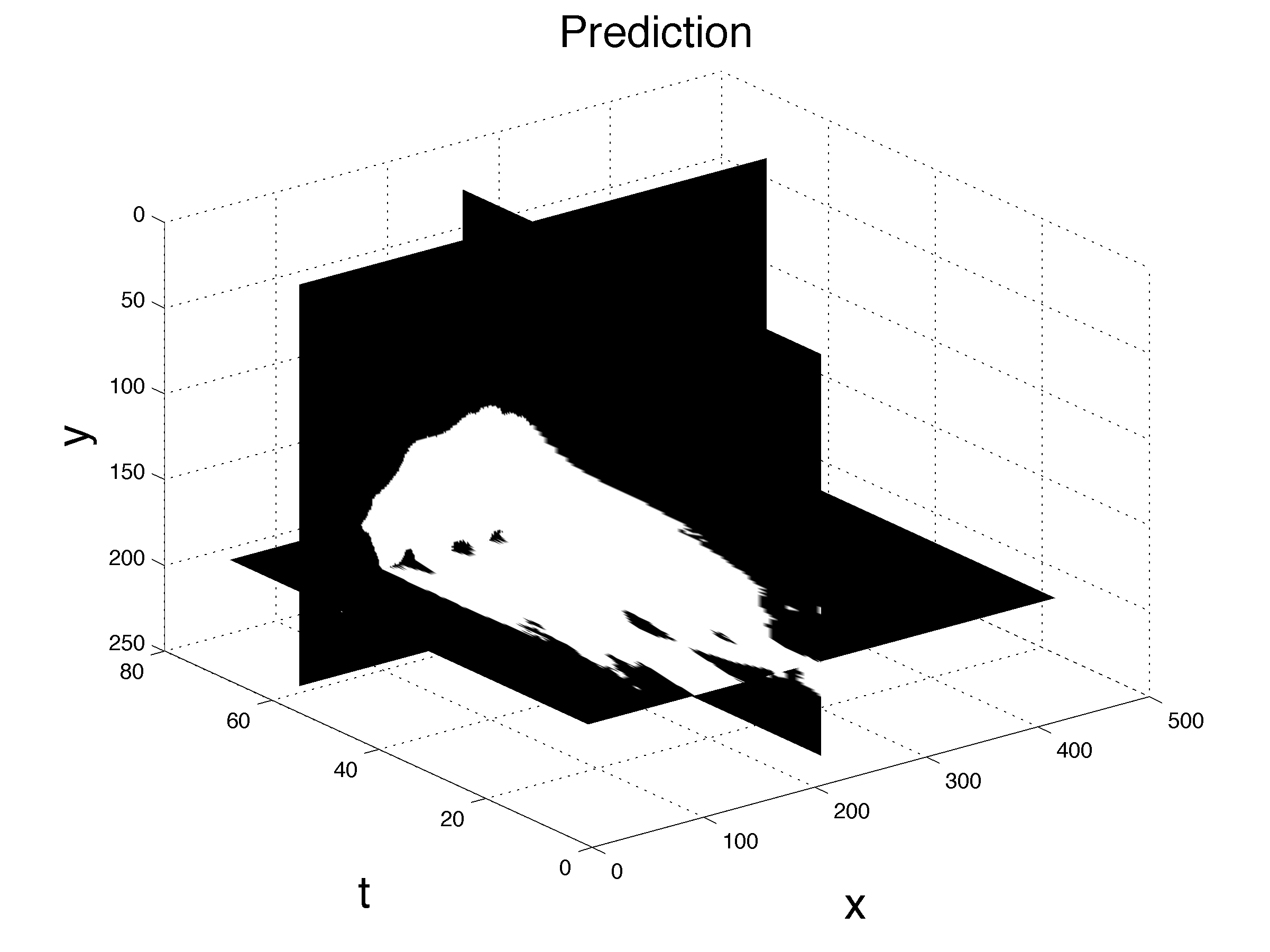}
        \label{fig:pred_mask}
    \end{subfigure}
    \caption{Three orthogonal slices through the data and prediction for the video-to-video training. Training only used 3 labeled time-slices. The training label consists of a 3D volume with just 3D annotated slices}
    \label{fig:BearVideo}
\end{center}
\end{figure*}

\subsubsection{Training a video-to-video network for single video segmentation}

When segmenting a video or other 3D/4D data, we hope to exploit the similarity between the time-frames and obtain better results compared to single image segmentation. The obvious challenge is memory: the cubic memory scaling with input size prohibits the storage of activations for deep and wide convolutional neural networks on a GPU, as required by automatic differentiation software.

Workarounds for the memory issues are often based on 2D slices to reduce the input size. Approaches range from slice-by-slice to sophisticated algorithms that combine neural-networks (encoder-decoder, GANs, recurrent networks) with other processing to arrive at hybrid methods that detect and propagate the segmentation in time, based on one or a few reference segmentation slices. See, e.g., \cite{Switch_10.1007/978-3-030-01234-2_6,IEEE8578868,caelles2019fast} for some recent examples that include literature review. 

Existing 3D CNNs are limited to low-resolution video or a small number of time-frames. \cite{hou2017end} use only eight time-frames at once, which limits the temporal information than can be exploited. \cite{Vox2Vox7789547} use $16$ frames starting at a resolution of $128 \times 171$.

We test our network and proposed methodology on the Davis video dataset \cite{IEEEDavisDataset}. The results are without pre-training the network, and no pre/post-processing was applied. The input size for our network is $240 \times 424 \times 72 \times 6$. We arrive at six channels by repeating the RGB channels twice. The output is at the same resolution. The network is the same as in the previous hyperspectral example. The labels for training consist of three time-slices: slice $5$, $30$, and $69$, see Figure \ref{fig:BearVideo}. The goal is to interpolate and extrapolate the three slices to obtain the the true label as in Figure \ref{fig:BearVideo}. Our approach is to train the network to map the full input video directly to the full segmentation, based on the three labeled slices and guided by the data. The prediction is shown in Figure \ref{fig:BearVideo}, and we achieve a mean IOU of $92.7$, with a positive IOU of $87.3$ and a negative IOU of $98.0 $.

\begin{figure}[htbp!]
\begin{center}
	\begin{subfigure}[b]{\linewidth}
		\centering
        \includegraphics[width=0.5\textwidth]{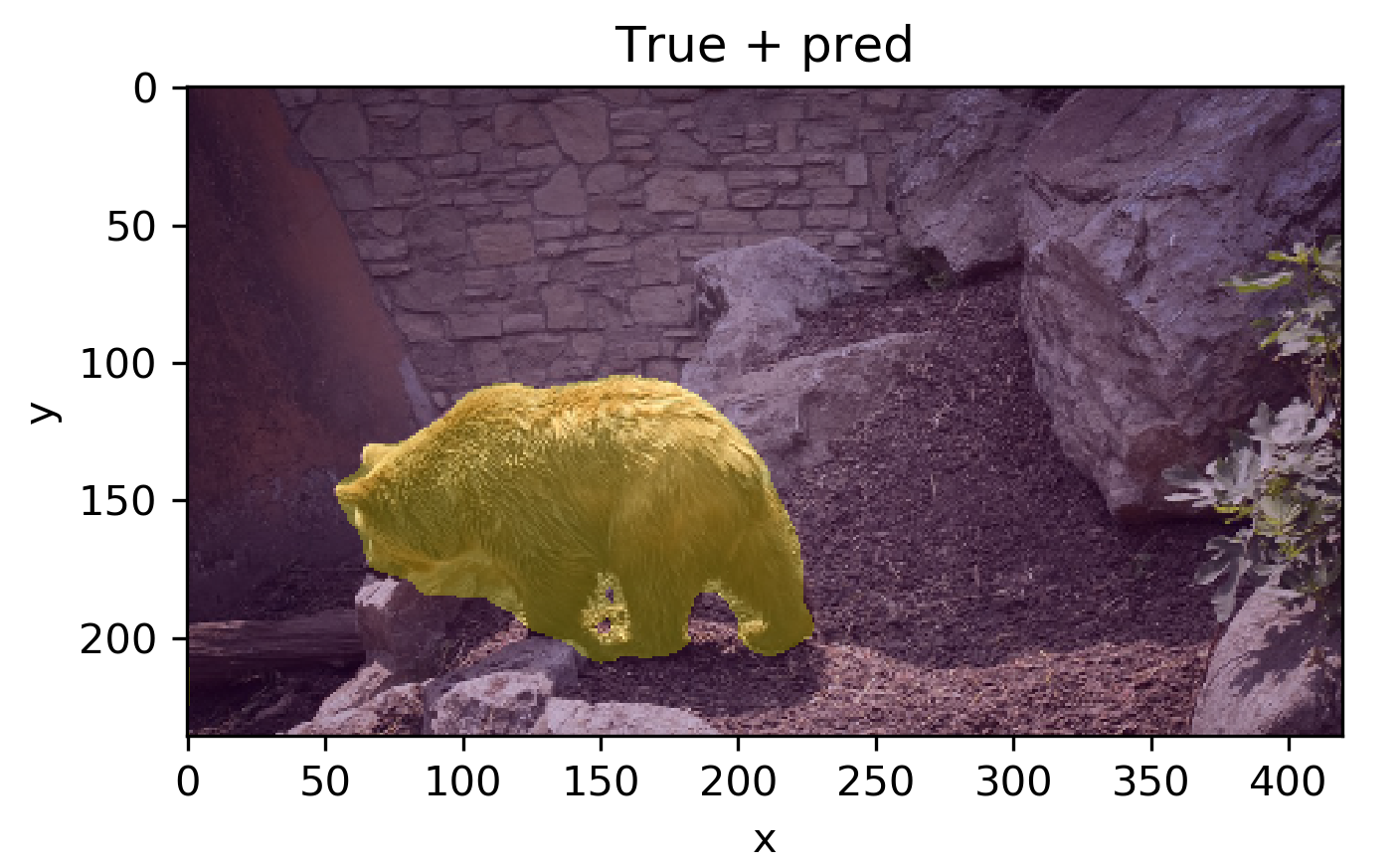}
        \label{fig:Bear_data_plus_segmentation}
	\end{subfigure}
\caption{A segmentation prediction from a single frame of the output of the video-to-segmentation network overlaid on the input video.}
\end{center}
\end{figure}

\subsection{Training a VAE}
\label{sec:vaegn_implementation}

In this subsection we discuss experiments that we have done in training a VAE on two data sets. First, we use the common MNIST data set \cite{lecun-mnisthandwrittendigit-2010} as a simple example and second, we have used the VAE in order to train a generative model based on the dataset Celeb-A \cite{liu2015faceattributes}. 
These networks are constructed as sets of consecutive layers where $\bfW = \bfI$ such that there is no pooling, followed by a layer where $\bfW$ takes the form of the 2D Haar wavelet described in Section \ref{buildingBlocks}. 
We call each of these sets a \textit{unit}, and the network structure is visualized in Figure \ref{network_diagram}. 

\subsubsection{Experiments with MNIST}

\begin{figure}[h]
\centering 
	\includegraphics[width=8cm]{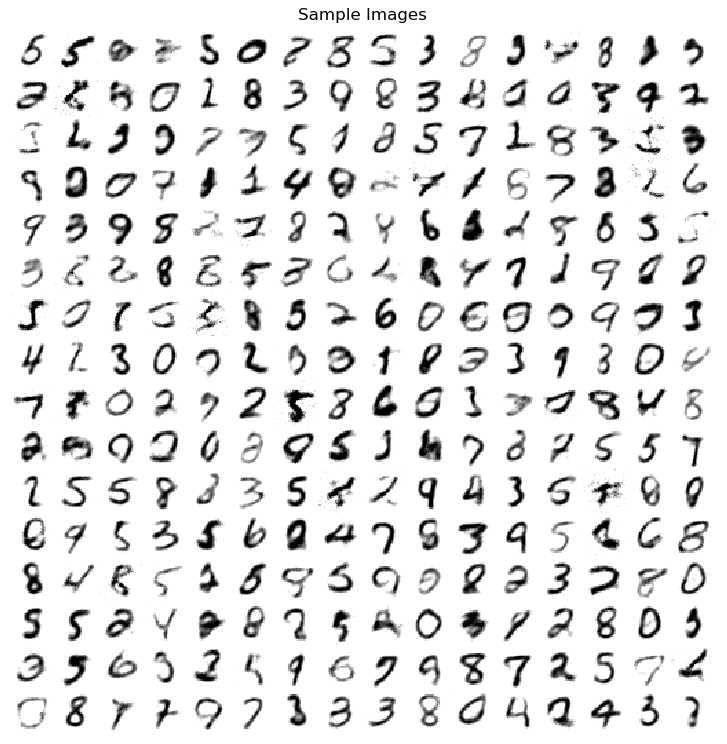}
	\caption{A sample of $256$ randomly sampled points from the latent space of the MNIST data set.}
\label{fig:mnist}
\end{figure}

We train our model on the MNIST images that have been resized to the nearest power of 2, $32\text{x}32$, and normalized using a channel mean of $0.1307$ and a channel standard deviation of $0.3081$.
We used a hyperbolic network with $5$ units, each with $3$ layers, and a step size $h=10^{-2}$. 
We train for a total of $100$ epochs using SGD with a batch size of $512$ and a learning rate of $10^{-2}$ for $\bftheta$ and $5 \times 10^{-3}$ for $\bfP$.
After training, we remove columns of $\bfP$ where the entries are less than $10^{-3}$, resulting in a $\bfP$ that is rectangular, reducing the latent space down to $8$ dimensions from the original $1024$.
We then randomly sample the latent sample with 256 random points, which resulted in the images displayed in Figure~\ref{fig:mnist}.
The majority of the images reasonably resemble images from the training set.
It is likely that the images that are non-plausible to the human eye correspond to parts of the latent space that exist on the boundary of two classes.
In a case as simple as MNIST, it is fair to assume that each class corresponds to a disjoint distribution in the image space.
Given that, when we normalize the latent space into a single continuous distribution it is fair to assume that there will be regions of that space that map to each class distribution.
Due to this fact, there must be interfaces in the latent space between each class, and sampled points from these boundaries will surely result in images that do not belong to the training image distribution.

\subsubsection{Experiments with Celeb-A}

In our next example we use the data set of faces in order to train a generative model of faces.

We used $20$ layer hyperbolic network with $6$ units $(4,4,4,4,2,2)$ for $100$ epochs using SGD with a learning rate of $10^{-1}$ and a batch size of 128.
We reduce our latent space down to $84$ dimensions from the original $12288$.

\begin{figure}[h]
\begin{center}

	\begin{subfigure}[t]{0.12\linewidth}
		\includegraphics[width=\textwidth]{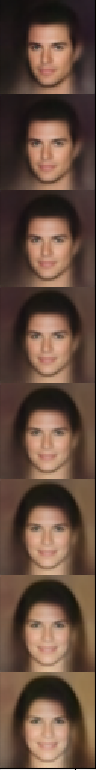}
		\label{fig:celeba_1}
		\caption{ }
	\end{subfigure}
	\begin{subfigure}[t]{0.12\linewidth}
		\includegraphics[width=\textwidth]{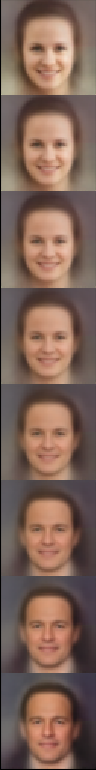}
		\label{fig:celeba_2}
		\caption{ }
	\end{subfigure}
	\begin{subfigure}[t]{0.12\linewidth}
		\includegraphics[width=\textwidth]{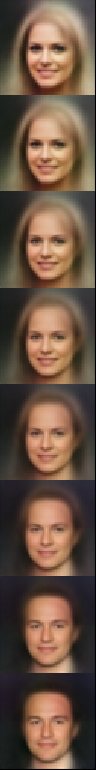}
		\label{fig:celeba_3}
		\caption{ }
	\end{subfigure}
	\begin{subfigure}[t]{0.12\linewidth}
		\includegraphics[width=\textwidth]{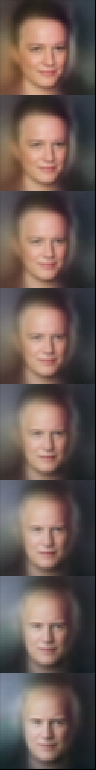}
		\label{fig:celeba_4}
		\caption{ }
	\end{subfigure}

\caption{Images resulting from interpolation across the latent space for CelebA dataset.}
\label{fig:celeba}
\end{center}
\end{figure}

In order to explore the learned manifold we select two points in the latent space from the validation set $\{z_a, z_b\} \in \bfZ$ and perform linear interpolation between them to get points $z_i$, $i \in \{ 0,1,...,N-1, N\}$ where $z_0 = z_a$ and $z_N = z_b$.
In Figure \ref{fig:celeba} we visualize $z_i$ for four pairs of points in the latent space along the columns. 
In each column, $z_a$ is presented in the first row, and $z_b$ is presented in the last row, with the linearly interpolated points $z_i$ in the middle rows.

We observe that the interpolation along the manifold exceeds results expected from a simple pixel wise interpolation. 
In all figures we see a smooth transition in facial features between each pair of faces, with all points in between being equally as plausible as the real points from the training set. 
We note that due to the high level of compression nearly all background information is lost, and only information about the structure of the face is maintained. 
In Figures \ref{fig:celeba}a, \ref{fig:celeba}b, \ref{fig:celeba}c we see smooth transitions in facial features across genders and hair colors. 
In Figure \ref{fig:celeba}d we see a smooth interpolation between faces of significantly different age.

\begin{figure}[h]
\begin{center}

	\includegraphics[width=0.7\textwidth]{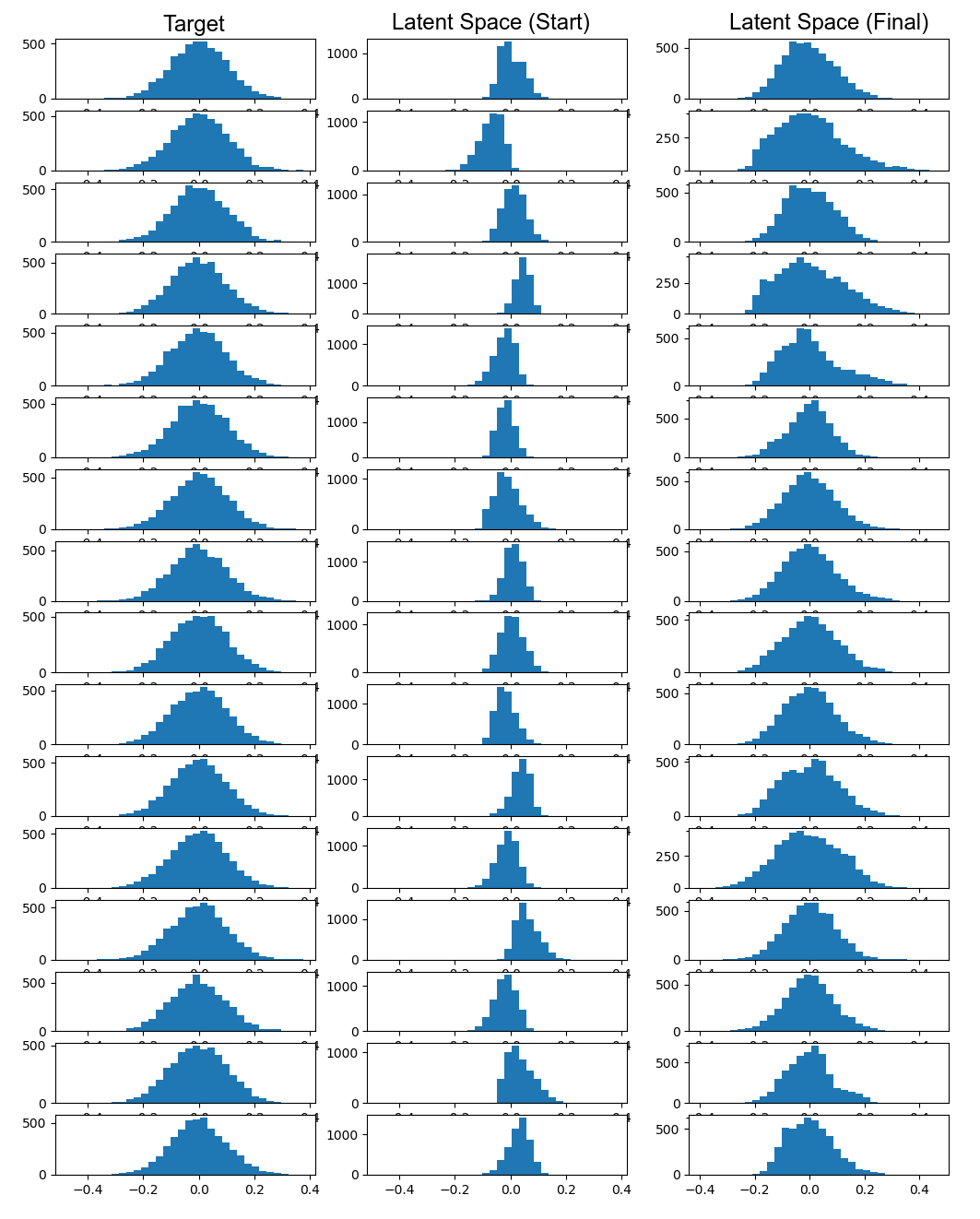}
	\caption{Projection of sample points onto 16 random directions for each space in order to qualitatively observe the normality of the latent space. The first column shows a sample from a Gaussian distribution. 
The second column shows a sample the complex latent space before training.
The third column shows a sample from the latent space after passing through the normalizing network. Note that each column contains 16 different projections for each space, and that there is no significance comparing across individual rows.}
	
\label{fig:celeba_vis}
\end{center}
\end{figure}

Testing for normality in high dimensions is notoriously difficult. 
Easily computed quantitative metrics such as mean, standard deviation, and skew provide some insight into the normality of a distribution, but these low mode statistics provide no guarantee especially in higher dimensions. 
We can visualize the normality of the latent spaces by visualizing marginals
projected onto random directions and visualize the histograms. 
In Figure \ref{fig:celeba_vis} we visualize the normality of the final latent space across $16$ random directions using a sample size of $1000$.
For comparison, we also visualize the target distribution $\mathcal{N}(0, 0.1)$.

We note that our network's latent space is reasonably normal compared to the target distribution. 
In some directions there are long tails and the distribution appears skewed, however in general the points appear to cluster into a normal distribution.
We acknowledge that visualizing these $16$ random directions represent no guarantee of normality.
However they do show that compared to the latent space of the auto encoder before training our network appears to be successful in obtaining a more normal distribution.

\section{Conclusions}
\label{conclusions}

In this work we have introduced a new architecture for deep neural networks. The architecture is motivated by the propagation of signals over physical networks where hyperbolic equations are used to describe the behavior of signal propagation. The network can be interpreted as a leapfrog discretization of the nonlinear Telegraph equation. The equation and its corresponding discretization are conservative, which implies that the network propagates the energy of the initial condition throughout all layers without decay. Similar to other networks, in order to obtain non-local behavior coarsening is used and, at the same time, the number of channels is increased. In order to coarsen the image conservatively, we use the discrete wavelet transform. Such a transform has a natural property that it increases the number of channels while reducing the resolution, while conserving all the information in the image due its invertibility. 
We have demonstrated the application of such networks on tasks such as segmentation as well as introduced a new application of the network for generative models using VAE's.

In our numerical experiments we show that despite the constraints introduced to enforce reversibility, the proposed hyperbolic network is able to succeed in tasks that commonly require large memory footprints, such as 3D video segmentation and hyperspectral image classification. We believe that for large scale problems and for problems in 3D, fully reversible networks will be a key in efficient training and inference.

\section*{Acknowledgments}
K.L and E.H acknowledge the support of the Natural Sciences and Engineering Research Council of Canada (NSERC).

\bibliographystyle{siamplain}
\bibliography{biblio, HyperSpectralRefs, vaegn}

\begin{thebibliography}{10}

\bibitem{AVENDI2016108}
{\sc M.~Avendi, A.~Kheradvar, and H.~Jafarkhani}, {\em A combined deep-learning
  and deformable-model approach to fully automatic segmentation of the left
  ventricle in cardiac mri}, Medical Image Analysis, 30 (2016), pp.~108 -- 119.

\bibitem{bengio2009learning}
{\sc Y.~Bengio}, {\em Learning deep architectures for {AI}}, Foundations and
  trends{\textregistered} in Machine Learning, 2 (2009), pp.~1--127.

\bibitem{brock2018large}
{\sc A.~Brock, J.~Donahue, and K.~Simonyan}, {\em Large scale gan training for
  high fidelity natural image synthesis}, 2018,
  \url{https://arxiv.org/abs/1809.11096}.

\bibitem{caelles2019fast}
{\sc S.~Caelles, A.~Pumarola, F.~Moreno-Noguer, A.~Sanfeliu, and L.~Van~Gool},
  {\em Fast video object segmentation with spatio-temporal gans}, arXiv
  preprint arXiv:1903.12161,  (2019).

\bibitem{Chang2017Reversible}
{\sc B.~Chang, L.~Meng, E.~Haber, L.~Ruthotto, D.~Begert, and E.~Holtham}, {\em
  Reversible architectures for arbitrarily deep residual neural networks}, in
  AAAI Conference on AI, 2018.

\bibitem{neuroODE}
{\sc T.~Q. Chen, Y.~Rubanova, J.~Bettencourt, and D.~K. Duvenaud}, {\em Neural
  ordinary differential equations}, CoRR, abs/1806.07366 (2018),
  \url{http://arxiv.org/abs/1806.07366},
  \url{https://arxiv.org/abs/1806.07366}.

\bibitem{Unet3D}
{\sc {\"O}.~{\c{C}}i{\c{c}}ek, A.~Abdulkadir, S.~S. Lienkamp, T.~Brox, and
  O.~Ronneberger}, {\em 3d u-net: Learning dense volumetric segmentation from
  sparse annotation}, in Medical Image Computing and Computer-Assisted
  Intervention -- MICCAI 2016, S.~Ourselin, L.~Joskowicz, M.~R. Sabuncu,
  G.~Unal, and W.~Wells, eds., Cham, 2016, Springer International Publishing,
  pp.~424--432.

\bibitem{dbOrthonormal}
{\sc I.~Daubechies}, {\em Orthonormal bases of compactly supported wavelets},
  Communication on Pure and Applied Mathematics, 41 (1988), pp.~909--996,
  \url{https://doi.org/10.1002/cpa.3160410705},
  \url{https://onlinelibrary.wiley.com/doi/abs/10.1002/cpa.3160410705},
  \url{https://arxiv.org/abs/https://onlinelibrary.wiley.com/doi/abs/10.1002/cpa.3160410705}.

\bibitem{DinhSB16}
{\sc L.~Dinh, J.~Sohl{-}Dickstein, and S.~Bengio}, {\em Density estimation
  using real {NVP}}, CoRR, abs/1605.08803 (2016),
  \url{http://arxiv.org/abs/1605.08803},
  \url{https://arxiv.org/abs/1605.08803}.

\bibitem{iUnet}
{\sc C.~Etmann, R.~Ke, and C.-B. Schönlieb}, {\em iunets: Fully invertible
  u-nets with learnable up- and downsampling}, 2020,
  \url{https://arxiv.org/abs/arXiv:2005.05220}.

\bibitem{fujieda2017wavelet}
{\sc S.~Fujieda, K.~Takayama, and T.~Hachisuka}, {\em Wavelet convolutional
  neural networks for texture classification}, arXiv preprint arXiv:1707.07394,
   (2017).

\bibitem{gholami2019anode}
{\sc A.~Gholami, K.~Keutzer, and G.~Biros}, {\em Anode: Unconditionally
  accurate memory-efficient gradients for neural odes}, 2019,
  \url{https://arxiv.org/abs/1902.10298}.

\bibitem{GomezEtAl2017}
{\sc A.~N. Gomez, M.~Ren, R.~Urtasun, and R.~B. Grosse}, {\em The reversible
  residual network: Backpropagation without storing activations}, in Adv Neural
  Inf Process Syst, 2017, pp.~2211--2221.

\bibitem{Goodfellow-et-al-2016}
{\sc I.~Goodfellow, Y.~Bengio, and A.~Courville}, {\em Deep Learning}, MIT
  Press, 2016.

\bibitem{Hammernik_2017}
{\sc K.~Hammernik, T.~Klatzer, E.~Kobler, M.~P. Recht, D.~K. Sodickson,
  T.~Pock, and F.~Knoll}, {\em Learning a variational network for
  reconstruction of accelerated mri data}, Magnetic Resonance in Medicine, 79
  (2017), p.~3055–3071, \url{https://doi.org/10.1002/mrm.26977},
  \url{http://dx.doi.org/10.1002/mrm.26977}.

\bibitem{Hanin2017DNNApprox}
{\sc B.~Hanin}, {\em Universal function approximation by deep neural nets with
  bounded width and relu activations}, arXiv preprint arXiv:1708.02691v3,
  (2017).

\bibitem{doi:10.1080/01431161.2018.1466079}
{\sc M.~Hasanlou and S.~T. Seydi}, {\em Hyperspectral change detection: an
  experimental comparative study}, International Journal of Remote Sensing, 39
  (2018), pp.~7029--7083, \url{https://doi.org/10.1080/01431161.2018.1466079},
  \url{https://doi.org/10.1080/01431161.2018.1466079},
  \url{https://arxiv.org/abs/https://doi.org/10.1080/01431161.2018.1466079}.

\bibitem{he2016deep}
{\sc K.~He, X.~Zhang, S.~Ren, and J.~Sun}, {\em Deep residual learning for
  image recognition}, in Proceedings of the IEEE Conference on Computer Vision
  and Pattern Recognition, 2016, pp.~770--778.

\bibitem{M8297014}
{\sc M.~{He}, B.~{Li}, and H.~{Chen}}, {\em Multi-scale 3d deep convolutional
  neural network for hyperspectral image classification}, in 2017 IEEE
  International Conference on Image Processing (ICIP), Sep. 2017,
  pp.~3904--3908, \url{https://doi.org/10.1109/ICIP.2017.8297014}.

\bibitem{hou2017end}
{\sc R.~Hou, C.~Chen, and M.~Shah}, {\em An end-to-end 3d convolutional neural
  network for action detection and segmentation in videos}, arXiv preprint
  arXiv:1712.01111,  (2017).

\bibitem{irevnet}
{\sc J.~Jacobsen, A.~W.~M. Smeulders, and E.~Oyallon}, {\em i-revnet: Deep
  invertible networks}, CoRR, abs/1802.07088 (2018),
  \url{http://arxiv.org/abs/1802.07088},
  \url{https://arxiv.org/abs/1802.07088}.

\bibitem{karras2017progressive}
{\sc T.~Karras, T.~Aila, S.~Laine, and J.~Lehtinen}, {\em Progressive growing
  of gans for improved quality, stability, and variation}, 2017,
  \url{https://arxiv.org/abs/1710.10196}.

\bibitem{karras2018stylebased}
{\sc T.~Karras, S.~Laine, and T.~Aila}, {\em A style-based generator
  architecture for generative adversarial networks}, 2018,
  \url{https://arxiv.org/abs/1812.04948}.

\bibitem{kingma2013autoencoding}
{\sc D.~P. Kingma and M.~Welling}, {\em Auto-encoding variational bayes}, 2013,
  \url{https://arxiv.org/abs/1312.6114}.

\bibitem{krizhevsky2012imagenet}
{\sc A.~Krizhevsky, I.~Sutskever, and G.~E. Hinton}, {\em Imagenet
  classification with deep convolutional neural networks}, in Advances in
  neural information processing systems, 2012, pp.~1097--1105.

\bibitem{lecun2015deep}
{\sc Y.~LeCun, Y.~Bengio, and G.~Hinton}, {\em Deep learning}, Nature, 521
  (2015), pp.~436--444.

\bibitem{lecun-mnisthandwrittendigit-2010}
{\sc Y.~LeCun and C.~Cortes}, {\em {MNIST} handwritten digit database}.
\newblock http://yann.lecun.com/exdb/mnist/, 2010,
  \url{http://yann.lecun.com/exdb/mnist/}.

\bibitem{C7729859}
{\sc H.~{Lee} and H.~{Kwon}}, {\em Contextual deep cnn based hyperspectral
  classification}, in 2016 IEEE International Geoscience and Remote Sensing
  Symposium (IGARSS), July 2016, pp.~3322--3325,
  \url{https://doi.org/10.1109/IGARSS.2016.7729859}.

\bibitem{doi:10.1080/2150704X.2019.1686780}
{\sc J.~Li, B.~Liang, and Y.~Wang}, {\em A hybrid neural network for
  hyperspectral image classification}, Remote Sensing Letters, 11 (2020),
  pp.~96--105, \url{https://doi.org/10.1080/2150704X.2019.1686780},
  \url{https://doi.org/10.1080/2150704X.2019.1686780},
  \url{https://arxiv.org/abs/https://doi.org/10.1080/2150704X.2019.1686780}.

\bibitem{rs9010067}
{\sc Y.~Li, H.~Zhang, and Q.~Shen}, {\em Spectral–spatial classification of
  hyperspectral imagery with 3d convolutional neural network}, Remote Sensing,
  9 (2017), \url{https://doi.org/10.3390/rs9010067},
  \url{https://www.mdpi.com/2072-4292/9/1/67}.

\bibitem{WaveletNetwork1}
{\sc P.~{Liu}, H.~{Zhang}, K.~{Zhang}, L.~{Lin}, and W.~{Zuo}}, {\em
  Multi-level wavelet-cnn for image restoration}, in 2018 IEEE/CVF Conference
  on Computer Vision and Pattern Recognition Workshops (CVPRW), June 2018,
  pp.~886--88609, \url{https://doi.org/10.1109/CVPRW.2018.00121}.

\bibitem{Switch_10.1007/978-3-030-01234-2_6}
{\sc S.~Liu, G.~Zhong, S.~De~Mello, J.~Gu, V.~Jampani, M.-H. Yang, and
  J.~Kautz}, {\em Switchable temporal propagation network}, in Computer Vision
  -- ECCV 2018, V.~Ferrari, M.~Hebert, C.~Sminchisescu, and Y.~Weiss, eds.,
  Cham, 2018, Springer International Publishing, pp.~89--104.

\bibitem{liu2015faceattributes}
{\sc Z.~Liu, P.~Luo, X.~Wang, and X.~Tang}, {\em Deep learning face attributes
  in the wild}, in Proceedings of International Conference on Computer Vision
  (ICCV), December 2015.

\bibitem{NIPS2016_6203}
{\sc W.~Luo, Y.~Li, R.~Urtasun, and R.~Zemel}, {\em Understanding the effective
  receptive field in deep convolutional neural networks}, in Advances in Neural
  Information Processing Systems 29, D.~D. Lee, M.~Sugiyama, U.~V. Luxburg,
  I.~Guyon, and R.~Garnett, eds., Curran Associates, Inc., 2016,
  pp.~4898--4906.

\bibitem{IEEE8578868}
{\sc S.~W. {Oh}, J.~{Lee}, K.~{Sunkavalli}, and S.~J. {Kim}}, {\em Fast video
  object segmentation by reference-guided mask propagation}, in 2018 IEEE/CVF
  Conference on Computer Vision and Pattern Recognition, June 2018,
  pp.~7376--7385, \url{https://doi.org/10.1109/CVPR.2018.00770}.

\bibitem{IEEEDavisDataset}
{\sc F.~{Perazzi}, J.~{Pont-Tuset}, B.~{McWilliams}, L.~V. {Gool}, M.~{Gross},
  and A.~{Sorkine-Hornung}}, {\em A benchmark dataset and evaluation
  methodology for video object segmentation}, in 2016 IEEE Conference on
  Computer Vision and Pattern Recognition (CVPR), June 2016, pp.~724--732,
  \url{https://doi.org/10.1109/CVPR.2016.85}.

\bibitem{doi:10.1190/INT-2018-0225.1}
{\sc B.~Peters, J.~Granek, and E.~Haber}, {\em Multiresolution neural networks
  for tracking seismic horizons from few training images}, Interpretation, 7
  (2019), pp.~SE201--SE213, \url{https://doi.org/10.1190/INT-2018-0225.1},
  \url{https://doi.org/10.1190/INT-2018-0225.1},
  \url{https://arxiv.org/abs/https://doi.org/10.1190/INT-2018-0225.1}.

\bibitem{ptvf}
{\sc W.~Press, S.~Teukolsky, W.~Vetterling, and B.~Flannery}, {\em Numerical
  Recipes in C: the Art of Scientific Computing}, Cambridge University Press,
  second~ed., 1992.

\bibitem{UNET2015}
{\sc O.~Ronneberger, P.~Fischer, and T.~Brox}, {\em U-net: Convolutional
  networks for biomedical image segmentation}, CoRR, abs/1505.04597 (2015),
  \url{http://arxiv.org/abs/1505.04597},
  \url{https://arxiv.org/abs/1505.04597}.

\bibitem{RuthottoHaber2018}
{\sc L.~Ruthotto and E.~Haber}, {\em Deep neural networks motivated by partial
  differential equations}, arXiv preprint arXiv:1804.04272,  (2018).

\bibitem{SalimansGZCRC16}
{\sc T.~Salimans, I.~J. Goodfellow, W.~Zaremba, V.~Cheung, A.~Radford, and
  X.~Chen}, {\em Improved techniques for training gans}, CoRR, abs/1606.03498
  (2016), \url{http://arxiv.org/abs/1606.03498},
  \url{https://arxiv.org/abs/1606.03498}.

\bibitem{SUnets}
{\sc S.~Shah, P.~Ghosh, L.~S. Davis, and T.~Goldstein}, {\em Stacked u-nets:
  {A} no-frills approach to natural image segmentation}, CoRR, abs/1804.10343
  (2018), \url{http://arxiv.org/abs/1804.10343},
  \url{https://arxiv.org/abs/1804.10343}.

\bibitem{Shelhamer2017FCN}
{\sc E.~Shelhamer, J.~Long, and T.~Darrell}, {\em Fully convolutional networks
  for semantic segmentation}, {IEEE} Trans. Pattern Anal. Mach. Intell., 39
  (2017), pp.~640--651, \url{https://doi.org/10.1109/TPAMI.2016.2572683},
  \url{https://doi.org/10.1109/TPAMI.2016.2572683}.

\bibitem{Szekely_testingfor}
{\sc G.~J. Székely and M.~L. Rizzo}, {\em Testing for equal distributions in
  high dimensions}, InterStat,  (2004).

\bibitem{Tao2018Deblurring}
{\sc X.~Tao, H.~Gao, Y.~Wang, X.~Shen, J.~Wang, and J.~Jia}, {\em
  Scale-recurrent network for deep image deblurring}, CoRR, abs/1802.01770
  (2018), \url{http://arxiv.org/abs/1802.01770},
  \url{https://arxiv.org/abs/1802.01770}.

\bibitem{Vox2Vox7789547}
{\sc D.~{Tran}, L.~{Bourdev}, R.~{Fergus}, L.~{Torresani}, and M.~{Paluri}},
  {\em Deep end2end voxel2voxel prediction}, in 2016 IEEE Conference on
  Computer Vision and Pattern Recognition Workshops (CVPRW), June 2016,
  pp.~402--409, \url{https://doi.org/10.1109/CVPRW.2016.57}.

\bibitem{WaveletReview}
{\sc F.~Truchetet and O.~Laligant}, {\em Wavelets in industrial applications: a
  review}, Wavelet Applications in Industrial Processing II, 5607 (2004),
  \url{https://doi.org/10.1117/12.580395},
  \url{https://doi.org/10.1117/12.580395}.

\bibitem{rs12010188}
{\sc Q.~Xu, Y.~Xiao, D.~Wang, and B.~Luo}, {\em Csa-mso3dcnn: Multiscale octave
  3d cnn with channel and spatial attention for hyperspectral image
  classification}, Remote Sensing, 12 (2020),
  \url{https://doi.org/10.3390/rs12010188},
  \url{https://www.mdpi.com/2072-4292/12/1/188}.

\bibitem{doi:10.1080/2150704X.2019.1681598}
{\sc Z.~Xue}, {\em A general generative adversarial capsule network for
  hyperspectral image spectral-spatial classification}, Remote Sensing Letters,
  11 (2020), pp.~19--28, \url{https://doi.org/10.1080/2150704X.2019.1681598},
  \url{https://doi.org/10.1080/2150704X.2019.1681598},
  \url{https://arxiv.org/abs/https://doi.org/10.1080/2150704X.2019.1681598}.

\bibitem{PSPNetZhao2017}
{\sc H.~Zhao, J.~Shi, X.~Qi, X.~Wang, and J.~Jia}, {\em Pyramid scene parsing
  network}, 2017 IEEE Conference on Computer Vision and Pattern Recognition
  (CVPR),  (2017), \url{https://doi.org/10.1109/cvpr.2017.660},
  \url{http://dx.doi.org/10.1109/CVPR.2017.660}.

\bibitem{Zhou2018Telegraph}
{\sc Y.~Zhou and Z.~Luo}, {\em A crank–nicolson collocation spectral method
  for the two-dimensional telegraph equations}, Journal of Inequalities and
  Applications, 2018:137 (2018),
  \url{https://journalofinequalitiesandapplications.springeropen.com/articles/10.1186/s13660-018-1728-5}.

\end{thebibliography}


\begin{thebibliography}{18}
\providecommand{\natexlab}[1]{#1}
\providecommand{\url}[1]{\texttt{#1}}
\expandafter\ifx\csname urlstyle\endcsname\relax
  \providecommand{\doi}[1]{doi: #1}\else
  \providecommand{\doi}{doi: \begingroup \urlstyle{rm}\Url}\fi

\bibitem[Barratt \& Sharma(2018)Barratt and Sharma]{barratt2018note}
Barratt, S. and Sharma, R.
\newblock A note on the inception score, 2018.

\bibitem[Brock et~al.(2018)Brock, Donahue, and Simonyan]{brock2018large}
Brock, A., Donahue, J., and Simonyan, K.
\newblock Large scale gan training for high fidelity natural image synthesis,
  2018.

\bibitem[Chen et~al.(2018)Chen, Rubanova, Bettencourt, and
  Duvenaud]{chen2018neural}
Chen, R. T.~Q., Rubanova, Y., Bettencourt, J., and Duvenaud, D.
\newblock Neural ordinary differential equations, 2018.

\bibitem[Dinh et~al.(2016)Dinh, Sohl{-}Dickstein, and Bengio]{DinhSB16}
Dinh, L., Sohl{-}Dickstein, J., and Bengio, S.
\newblock Density estimation using real {NVP}.
\newblock \emph{CoRR}, abs/1605.08803, 2016.
\newblock URL \url{http://arxiv.org/abs/1605.08803}.

\bibitem[Gholami et~al.(2019)Gholami, Keutzer, and Biros]{gholami2019anode}
Gholami, A., Keutzer, K., and Biros, G.
\newblock Anode: Unconditionally accurate memory-efficient gradients for neural
  odes, 2019.

\bibitem[Haber \& Ruthotto(2017)Haber and Ruthotto]{Haber_2017}
Haber, E. and Ruthotto, L.
\newblock Stable architectures for deep neural networks.
\newblock \emph{Inverse Problems}, 34\penalty0 (1):\penalty0 014004, Dec 2017.
\newblock ISSN 1361-6420.
\newblock \doi{10.1088/1361-6420/aa9a90}.
\newblock URL \url{http://dx.doi.org/10.1088/1361-6420/aa9a90}.

\bibitem[Karras et~al.(2017)Karras, Aila, Laine, and
  Lehtinen]{karras2017progressive}
Karras, T., Aila, T., Laine, S., and Lehtinen, J.
\newblock Progressive growing of gans for improved quality, stability, and
  variation, 2017.

\bibitem[Karras et~al.(2018)Karras, Laine, and Aila]{karras2018stylebased}
Karras, T., Laine, S., and Aila, T.
\newblock A style-based generator architecture for generative adversarial
  networks, 2018.

\bibitem[Kingma \& Dhariwal(2018)Kingma and Dhariwal]{kingma2018glow}
Kingma, D.~P. and Dhariwal, P.
\newblock Glow: Generative flow with invertible 1x1 convolutions, 2018.

\bibitem[Kingma \& Welling(2013)Kingma and Welling]{kingma2013autoencoding}
Kingma, D.~P. and Welling, M.
\newblock Auto-encoding variational bayes, 2013.

\bibitem[Kobyzev et~al.(2019)Kobyzev, Prince, and
  Brubaker]{kobyzev2019normalizing}
Kobyzev, I., Prince, S., and Brubaker, M.~A.
\newblock Normalizing flows: An introduction and review of current methods,
  2019.

\bibitem[LeCun \& Cortes(2010)LeCun and
  Cortes]{lecun-mnisthandwrittendigit-2010}
LeCun, Y. and Cortes, C.
\newblock {MNIST} handwritten digit database.
\newblock http://yann.lecun.com/exdb/mnist/, 2010.
\newblock URL \url{http://yann.lecun.com/exdb/mnist/}.

\bibitem[Lensink et~al.(2019)Lensink, Haber, and Peters]{lensink2019fully}
Lensink, K., Haber, E., and Peters, B.
\newblock Fully hyperbolic convolutional neural networks, 2019.

\bibitem[Liu et~al.(2015)Liu, Luo, Wang, and Tang]{liu2015faceattributes}
Liu, Z., Luo, P., Wang, X., and Tang, X.
\newblock Deep learning face attributes in the wild.
\newblock In \emph{Proceedings of International Conference on Computer Vision
  (ICCV)}, December 2015.

\bibitem[Rodriguez(2013)]{tvreg}
Rodriguez, P.
\newblock Total variation regularization algorithms for images corrupted with
  different noise models: A review.
\newblock \emph{Journal of Electrical and Computer Engineering}, 2013, 07 2013.
\newblock \doi{10.1155/2013/217021}.

\bibitem[Salimans et~al.(2016)Salimans, Goodfellow, Zaremba, Cheung, Radford,
  and Chen]{SalimansGZCRC16}
Salimans, T., Goodfellow, I.~J., Zaremba, W., Cheung, V., Radford, A., and
  Chen, X.
\newblock Improved techniques for training gans.
\newblock \emph{CoRR}, abs/1606.03498, 2016.
\newblock URL \url{http://arxiv.org/abs/1606.03498}.

\bibitem[Székely \& Rizzo(2004)Székely and Rizzo]{Szekely_testingfor}
Székely, G.~J. and Rizzo, M.~L.
\newblock Testing for equal distributions in high dimensions.
\newblock \emph{InterStat}, 2004.

\bibitem[Zhu et~al.(2017)Zhu, Park, Isola, and Efros]{zhu2017unpaired}
Zhu, J.-Y., Park, T., Isola, P., and Efros, A.~A.
\newblock Unpaired image-to-image translation using cycle-consistent
  adversarial networks, 2017.

\end{thebibliography}

\end{document}